\documentclass[10pt,twocolumn,letterpaper]{article}

\usepackage{cvpr}              
\definecolor{cvprblue}{rgb}{0.21,0.49,0.74}
\usepackage[pagebackref,breaklinks,colorlinks,allcolors=cvprblue]{hyperref}

\def\paperID{*****} 
\def\confName{CVPR}
\def\confYear{2026}

\usepackage{booktabs}
\usepackage{xcolor}
\usepackage{pifont}
\usepackage{makecell}
\usepackage{marvosym} 

\definecolor{mygreen}{RGB}{0,120,0}
\definecolor{myred}{RGB}{180,40,40}

\newcommand{\cmark}{\textcolor{mygreen}{\ding{51}}}
\newcommand{\xmark}{\textcolor{myred}{\ding{55}}}

\usepackage{marvosym} 

\title{PortraitCraft: A Benchmark for Portrait Composition Understanding and Generation}

\author{Yuyang Sha\thanks{Equal contribution} \\
MT Lab, Meitu Inc\\
{\tt\small syy5@meitu.com}
\and
Zijie Lou\footnotemark[1] \\
MT Lab, Meitu Inc\\
{\tt\small lzj7@meitu.com}
\and
Youyun Tang\footnotemark[1] \\
MT Lab, Meitu Inc\\
{\tt\small tyy1@meitu.com}
\and
Xiaochao Qu\\
MT Lab, Meitu Inc\\
{\tt\small qxc@meitu.com}
\and
Zheng Qu\\
Airbrush Studio, Meitu Inc\\
{\tt\small lia.qu@pixocial.com}
\and
Ben Xia\\
Meitu Design, Meitu Inc\\
{\tt\small ben@meitu.com}
\and
Haoxiang Li\\
MT Lab, Meitu Inc\\
{\tt\small haoxiang.li@pixocial.com}
\and
Ting Liu \textsuperscript{\Letter} \thanks{Project leader} \\
MT Lab, Meitu Inc\\
{\tt\small lt@meitu.com}
\and
Luoqi Liu \textsuperscript{\Letter} \\
MT Lab, Meitu Inc\\
{\tt\small llq5@meitu.com}
}

\begin{document}
\maketitle
\begin{abstract}

Portrait composition plays a central role in portrait aesthetics and visual communication, yet existing datasets and benchmarks mainly focus on coarse aesthetic scoring, generic image aesthetics, or unconstrained portrait generation. This limits systematic research on structured portrait composition analysis and controllable portrait generation under explicit composition requirements. In this paper, we introduce PortraitCraft, a unified benchmark for portrait composition understanding and generation. PortraitCraft is built on a dataset of approximately 50,000 curated real portrait images with structured multi-level supervision, including global composition scores, annotations over 13 composition attributes, attribute-level explanation texts, visual question answering pairs, and composition-oriented textual descriptions for generation. Based on this dataset, we establish two complementary benchmark tasks for composition understanding and composition-aware generation within a unified framework. The first evaluates portrait composition understanding through score prediction, fine-grained attribute reasoning, and image-grounded visual question answering, while the second evaluates portrait generation from structured composition descriptions under explicit composition constraints. We further define standardized evaluation protocols and provide reference baseline results with representative multimodal models. PortraitCraft provides a comprehensive benchmark for future research on fine-grained portrait understanding, interpretable aesthetic assessment, and controllable portrait generation.

\end{abstract}    
\section{Introduction}
\label{sec:intro}

Portrait images play an important role in visual content creation and are widely used in photography, advertising, social media production, image editing, and AI-generated images~\cite{piq23,c-1}. In these applications, image quality is not determined only by resolution or realism but also by compositional quality. Important factors include clear subject emphasis, appropriate cropping, and natural body pose together with facial expression~\cite{finecaption}. The interaction between the subject and the background further affects visual quality, while the overall visual atmosphere influences how portrait images are perceived and evaluated. As a result, portrait composition has become an important problem in visual aesthetics, creative image understanding, and controllable image generation. Modeling portrait composition is challenging because portrait images involve strong subject-centered constraints and multiple interacting factors~\cite{artimuse,para}. Compared with generic image aesthetics, portrait composition depends not only on technical quality and visual hierarchy, but also on body pose, facial expression, scene context, emotional tone, and overall artistic coherence~\cite{c-2}. These factors are closely connected and cannot be adequately captured by a single global score. In practice, portrait composition assessment requires more than judging whether an image has high visual quality. It also requires explaining why the composition is effective or ineffective. This requirement becomes even more important in image generation, where models are expected to produce portraits that are both visually plausible and compositionally appropriate~\cite{c-3}.

Recent progress in image aesthetics assessment has shown the growing importance of structured and fine-grained evaluation~\cite{ase-1}. Rather than relying only on coarse global scores, recent studies increasingly emphasize multi-attribute analysis, richer annotation schemes, and more interpretable assessment. This shift is especially important for portrait composition, where visual quality depends on multiple interrelated factors and cannot be adequately captured by a single score alone. Despite this progress, existing public datasets still do not adequately support portrait composition research. Classical datasets such as AVA~\cite{ava} and AADB~\cite{aadb} have advanced generic image aesthetic assessment, but they were not designed specifically for portrait composition. Their annotations are also limited in supporting portrait-centered, fine-grained, and interpretable evaluation. Datasets such as APDDv2~\cite{apddv2} provide richer annotations in broader artistic domains, yet their primary focus is paintings and drawings rather than portrait photography. More recent resources~\cite{artimuse,unireal} reflect growing interest in detailed portrait quality perception and fine-grained visual judgment. However, they still do not provide a unified large-scale benchmark that jointly supports portrait composition understanding and composition-aware generation. In addition, public resources with precise annotations and high-quality expert evaluation remain limited. As a result, a large-scale portrait-centered benchmark with strong annotation quality and standardized evaluation is still lacking.

\begin{figure*}[t]
    \centering
    \includegraphics[width=\textwidth]{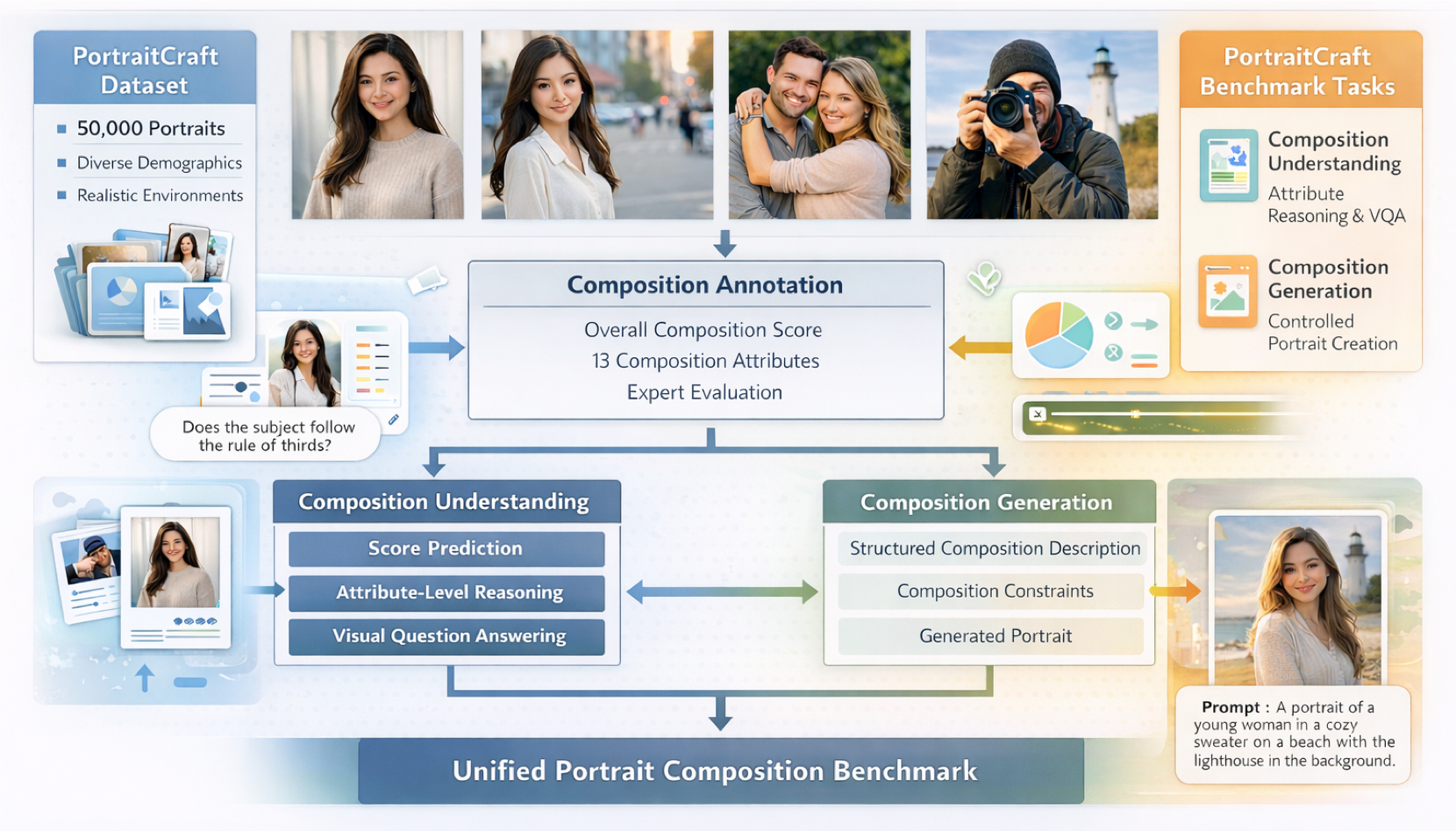}
    \caption{
    Overview of the PortraitCraft benchmark. PortraitCraft is built on 50,000 curated real portrait images and provides a unified framework for portrait composition understanding and generation. The benchmark includes two related tasks. Track 1 evaluates portrait composition understanding through overall score prediction, fine-grained attribute-level analysis, and image-based visual question answering. Track 2 evaluates portrait composition generation from structured composition descriptions under explicit composition constraints. Together, these two tasks connect composition analysis with composition-guided generation and support structured research on portrait composition.
    }
    \label{fig:overview}
\end{figure*}
 
To address this gap, we introduce PortraitCraft, a benchmark for portrait composition understanding and generation. An overview of the PortraitCraft benchmark is illustrated in Fig.~\ref{fig:overview}. PortraitCraft focuses on images in which humans are the primary visual subjects and covers diverse portrait scenarios, including single-person and multi-person images, as well as full-body and half-body compositions. The benchmark is organized around two closely related tasks. The first task, portrait composition understanding, evaluates whether models can assess composition quality in a structured way. It requires models to predict overall composition scores, make fine-grained attribute judgments, and answer image-based visual questions. The second task, portrait composition generation, evaluates whether models can generate portrait images that match structured composition descriptions. It focuses on generation under explicit composition constraints rather than conventional text-to-image synthesis. Through this design, these two tasks establish a unified benchmark that connects composition understanding, composition representation, and portrait composition generation.

A key strength of PortraitCraft lies in its data construction pipeline and annotation design. Rather than relying on weak labels or crowd-sourced annotations, PortraitCraft is built through a combination of model-based pre-screening and expert evaluation. Multiple professional designers participated in the assessment process, bringing extensive experience in visual design and aesthetic evaluation. Their expertise supports reliable aesthetic judgment and consistent annotation quality. As a result, PortraitCraft provides carefully curated portrait images together with structured supervision for both benchmark tasks, including overall composition scores, fine-grained attribute labels, image-based visual questions, and composition-oriented textual descriptions. This design enables PortraitCraft to support both multi-level portrait composition understanding and portrait composition generation guided by explicit structural constraints. Beyond the dataset itself, PortraitCraft also emphasizes standardized evaluation. We define task-specific evaluation protocols for both benchmark tasks to support consistent and reproducible comparison. For portrait composition understanding, the evaluation covers overall score prediction, fine-grained attribute judgments, and image-based visual question answering. For portrait composition generation, the evaluation focuses on alignment between generated images and target composition specifications, rather than conventional image generation metrics alone. Therefore, PortraitCraft formulates portrait composition as a structured benchmark problem with explicit criteria, standardized testing procedures, and baseline results. We hope it will support future research on portrait composition understanding, structured composition modeling, and controllable portrait composition generation.

Our contributions can be summarized as follows:
\begin{itemize}
  \item We introduce \textbf{PortraitCraft}, a unified benchmark for portrait composition understanding and generation that integrates global composition assessment, fine-grained attribute reasoning, and composition-aware portrait generation within a single evaluation framework.
  
  \item We construct a large-scale portrait composition dataset about 50,000 curated real images and structured multi-level supervision, including global composition scores, annotations over 13 composition attributes, attribute-level explanation texts, visual question answering pairs, and composition-oriented textual descriptions for controllable generation. The dataset is publicly available at \url{https://huggingface.co/datasets/zijielou/PortraitCraft}.

  \item We establish standardized evaluation protocols for both benchmark tracks and provide reference baseline results with representative multimodal models, enabling reproducible and comparable evaluation for portrait composition understanding and generation. The reference implementation is available at \url{https://github.com/yytang25/qwen-vl-ft}.

\end{itemize}

\section{Related Work}
\label{sec:related_work}

\subsection{Aesthetic and Composition Datasets}
Existing datasets related to aesthetics and composition have supported important progress in image quality assessment and visual preference analysis. Early benchmarks such as AVA~\cite{ava} and AADB~\cite{aadb} mainly support generic image-level aesthetic scoring. More recent datasets introduce richer attribute annotations and focus on more specialized domains, including artistic images and portrait quality assessment, as represented by APDDv2~\cite{apddv2}, PIQ23~\cite{piq23}, and LAPIS~\cite{lapis}. These efforts reflect a clear shift from global aesthetic scoring toward more structured and fine-grained evaluation. Despite these advances, existing datasets remain insufficient for portrait composition research. Most are not designed around portrait-centered composition and do not provide supervision that jointly supports composition scoring, attribute-level reasoning, and composition-aware generation~\cite{c-8}. Moreover, aesthetic annotation is inherently difficult to standardize and often relies on limited labeling protocols, while expert-driven evaluation remains uncommon in large-scale public datasets. As a result, current resources do not yet provide a unified benchmark for structured portrait composition understanding and generation.

\subsection{Portrait Composition Understanding and Generation}
Research on portrait composition understanding is closely related to image aesthetic assessment and composition quality analysis~\cite{c-4}. Early studies mainly focused on predicting global aesthetic scores using datasets such as AVA~\cite{ava}, while more recent work explored attribute-level evaluation and more structured aesthetic assessment. However, these approaches are typically designed for generic image aesthetics rather than portrait-centered composition, and they rarely support a unified setting that jointly requires composition scoring, fine-grained attribute judgments, and image-grounded reasoning. Portrait composition generation is related to portrait synthesis and controllable image generation. Existing methods often rely on semantic descriptions, facial attributes, or layout constraints to guide image generation~\cite{c-5}. However, they mainly focus on visual realism, identity consistency, or semantic alignment, rather than explicit composition modeling. As a result, portrait composition understanding and portrait generation are typically studied independently, and a unified benchmark connecting structured composition understanding with composition-guided portrait generation remains lacking~\cite{c-6}.

\subsection{Multimodal Large Language Models}
Recent multimodal large language models (MLLMs) have shown strong performance in visual understanding and multimodal reasoning~\cite{kacl,qwen3vl,llava}. They have been widely applied to tasks such as image description, visual question answering, and fine-grained visual analysis~\cite{vlm,mdd-llm,mdd-thinker}. These advances make MLLMs highly relevant to portrait composition research, especially for problems that require global quality judgment, attribute-level analysis, and image-grounded reasoning~\cite{c-7}. Their ability to process structured textual input also creates new possibilities for composition-guided image generation. However, current evaluations of MLLMs are still largely centered on general-purpose multimodal benchmarks. As a result, their capability in portrait composition understanding and generation has not been systematically studied. This limitation motivates the development of PortraitCraft, which supports targeted evaluation of multimodal models on both portrait composition understanding and portrait composition generation. 

\section{PortraitCraft Benchmark}
\label{sec:benchmark}

\subsection{Benchmark Overview}
PortraitCraft is a benchmark for portrait composition understanding and generation. Portrait composition is commonly studied through global aesthetic scoring or within general image generation settings. PortraitCraft instead formulates portrait composition as a structured benchmark problem. The goal is to evaluate whether models can both analyze composition quality in portrait images and generate portraits that satisfy explicit composition requirements. In this way, the benchmark provides a unified setting for studying portrait composition as both a visual understanding problem and a composition-aware generation problem. To support this objective, the PortraitCraft benchmark includes two closely related tasks: portrait composition understanding and portrait composition generation. The first task evaluates whether models can perform structured composition analysis through overall score prediction, fine-grained attribute judgments, and image-based visual question answering. The second task evaluates whether models can generate portrait images from structured composition descriptions under explicit compositional constraints. Although these two tasks are conceptually connected and share a common representation space of portrait composition, they can be studied independently or jointly depending on the research setting. The two tasks jointly define a flexible benchmark framework that supports research on portrait composition from analysis to generation.

\begin{table*}[t]
\centering
\footnotesize
\setlength{\tabcolsep}{5pt}
\begin{tabular}{l l r c l l c c}
\toprule
\textbf{Dataset} & \textbf{Domain} & \textbf{Scale} & \textbf{Expert} & \textbf{Annotation} & \textbf{Attributes} & \textbf{Understanding} & \textbf{Generation} \\
\midrule
AVA~\cite{ava} & Generic Aesthetics & 255,528 & \xmark & Overall Score & None & Limited & \xmark \\
Flickr-AES~\cite{flickr} & Generic Aesthetics & 40,499 & \xmark & Overall Score & None & Limited & \xmark \\
AADB~\cite{aadb} & Generic Aesthetics & 10,000 & \xmark & Overall Score, Attributes & Limited & \cmark & \xmark \\
ArtiMuse~\cite{artimuse} & Generic Aesthetics & 10,000 & \cmark & Overall Score, Attributes & 8 Attributes & \cmark & \xmark \\
APDDv2~\cite{apddv2} & Artistic Images & 10,023 & \cmark & \makecell[l]{Overall Score, Attributes, \\ Comments} & 10 Attributes & \cmark & \xmark \\
PIQ23~\cite{piq23} & Portrait Quality & 5,116 & \cmark & Quality Labels & 3 Attributes & Limited & \xmark \\
PARA~\cite{para} & Portrait Aesthetics & 31,299 & \cmark & Ranking Labels & Limited & Limited & \xmark \\ 
\midrule
\textbf{PortraitCraft} & \textbf{Portrait Composition} & \textbf{49,863} & \cmark & \makecell[l]{\textbf{Overall Score, Attributes,} \\ \textbf{VQA, Structured Descriptions}} & \makecell[l]{\textbf{13 Composition} \\ \textbf{Attributes}} & \cmark & \cmark \\
\bottomrule
\end{tabular}
\caption{Comparison of PortraitCraft with representative aesthetic and composition-related datasets. \textbf{Understanding} indicates whether the dataset supports structured composition analysis beyond coarse overall scoring.}
\label{tab:t1}
\end{table*}

\subsection{Track 1: Portrait Composition Understanding}
Track 1 evaluates portrait composition understanding as a multi-level visual task rather than a single-score prediction problem. Given a portrait image, the model is required to assess composition quality at different levels of granularity. The task contains three components: overall composition score prediction, fine-grained judgments over 13 composition attributes, and image-based multiple-choice visual question answering. These three components evaluate global composition judgment, structured attribute-level analysis, and fine-grained image-grounded reasoning, respectively.  

The training data for Track 1 are provided in the form of image-text pairs. Each training sample consists of a portrait image and a corresponding text description. The text includes an overall composition score, the scores of 13 composition attributes, and explanations of these attribute-level judgments. This design provides structured supervision together with explanation-oriented guidance. It allows the model to learn both the outcomes of composition assessment and the reasoning patterns behind fine-grained judgments. At test time, the input is a single portrait image. The model is required to answer three questions. First, it predicts an overall composition score for the image. Second, it assigns one of three levels, good, medium, or poor, to each of the 13 composition attributes. At this stage, the model is not required to explain the reasons behind these judgments. Third, it answers one image-based multiple-choice question with four highly distractive options. This testing setup requires the model to demonstrate a deep understanding of both overall composition quality and fine-grained visual details.

The 13 attributes cover several important aspects of portrait composition. These aspects include color and style, technical and lighting quality, composition principles and spatial organization, as well as subject presentation and visual stability. They collectively define a structured attribute space for portrait composition analysis. Track 1 defines a structured representation space for portrait composition and provides the foundation for composition understanding in the PortraitCraft benchmark.

\subsection{Track 2: Portrait Composition Generation}
Track 2 evaluates portrait composition generation under structured compositional constraints. Unlike conventional text-to-image generation, this task does not focus on free-form image synthesis from open-ended prompts. Instead, it requires the model to generate a portrait image that matches a given composition-oriented textual description. The goal is to test whether the model can understand explicit composition requirements and translate them into a coherent portrait image.

The training data for Track 2 consist of portrait images paired with structured composition-oriented textual descriptions. These descriptions encode composition-related requirements such as subject placement, spatial organization, visual center, negative space, depth layering, and other relevant structural cues. Through these image-text pairs, the model learns the mapping from structured composition descriptions to portrait images. At test time, the input is a single textual description only, without any reference image. The model is required to generate a portrait image that satisfies the composition specifications expressed in the text.

The evaluation of Track 2 is centered on composition alignment rather than conventional image generation quality. The benchmark does not primarily assess whether the generated image appears highly realistic or visually similar to a reference image. Instead, it focuses on whether the generated result follows the composition rules and fine-grained structural requirements described in the input text. To support this evaluation, submitted images are analyzed by an internal model that extracts structured composition properties from the generated results. These parsed properties are then compared with the target textual specifications to compute the final score. In this way, Track 2 serves as the generation component of PortraitCraft and evaluates whether portrait composition knowledge can be translated into controllable image generation.

\subsection{Task Relationship}
The two tasks in PortraitCraft play complementary roles in the study of portrait composition. Track 1 focuses on structured composition understanding by requiring models to assess overall composition quality, make fine-grained judgments over multiple composition attributes, and perform image-grounded reasoning. In this way, it defines portrait composition through global evaluation, attribute-level analysis, and fine-grained visual understanding. Track 2 focuses on generation under explicit composition constraints. It examines whether structured composition knowledge can be translated into image generation that follows target composition requirements. The two tasks therefore address portrait composition from two connected perspectives. One emphasizes how composition should be analyzed and represented, while the other tests whether such knowledge can be executed in generation. Each task can be studied independently, but considered together they provide a broader and more complete view of portrait composition.

This dual-task design also has broader value for both research and practice. From a research perspective, it connects composition understanding, representation, and generation within a single benchmark, making it possible to study how structured composition knowledge can move from analysis to controllable synthesis. From a practical perspective, real-world portrait systems often require both capabilities. They need to assess composition quality in existing images and generate or refine images according to composition-related goals. By combining these two complementary tasks in one benchmark, PortraitCraft provides a more complete and flexible framework for advancing portrait composition research.

\section{PortraitCraft Dataset}
\label{sec:dataset}

\subsection{Dataset Overview}
PortraitCraft is a portrait-centered dataset for portrait composition understanding and portrait composition generation. It contains about 50,000 curated portrait images and organizes supervision around a structured representation of portrait composition. The dataset includes overall composition scores, fine-grained composition attributes, image-based visual question answering supervision, and structured composition descriptions for generation. Table~\ref{tab:t1} compares PortraitCraft with representative aesthetic and composition-related datasets. Compared with existing resources, PortraitCraft provides several clear advantages. First, it is explicitly designed for portrait composition rather than generic aesthetics, artistic images, or portrait quality alone. Second, it combines expert-involved evaluation with richer supervision, including overall scores, 13 composition attributes, VQA, and structured composition descriptions. Third, it supports both composition understanding and composition-aware generation within a unified dataset. These properties make PortraitCraft a more comprehensive benchmark resource for structured portrait composition research.

\subsection{Image Curation}
The images in PortraitCraft were collected from Unsplash, a public photography platform that provides high-resolution images with diverse visual content and permissive usage conditions for academic research. Compared with images from social media sources, Unsplash photographs typically present clearer visual structure and fewer privacy concerns, making them suitable for portrait composition analysis. Using keyword-based retrieval, category filtering, and preliminary visual model screening, we constructed an initial candidate pool of approximately 100,000 images related to human subjects. The candidate set covers a wide range of portrait scenarios, including single-person and multi-person images, half-body and full-body portraits, as well as everyday human-centered scenes in natural environments. To ensure suitability for portrait composition research, we applied a multi-stage filtering process combining automatic screening and manual inspection. Images were removed if they exhibited severe blur, low resolution, extreme exposure problems, weak subject prominence, invalid composition structure, or non-photographic content. In addition, images that were likely to be AI-generated were filtered through both model-based detection and manual verification to ensure that the dataset contains only real photographs. After this curation process, about 50,000 high-quality portrait images were retained as the final dataset. During selection, we also considered diversity across portrait configurations, including single-person and multi-person scenes, half-body and full-body representations, and different composition styles. The filtering procedure combined automated selection with multiple rounds of manual review to improve consistency and reliability of the curated dataset.

\begin{figure}[t]
\centering
\includegraphics[width=1.0\linewidth]{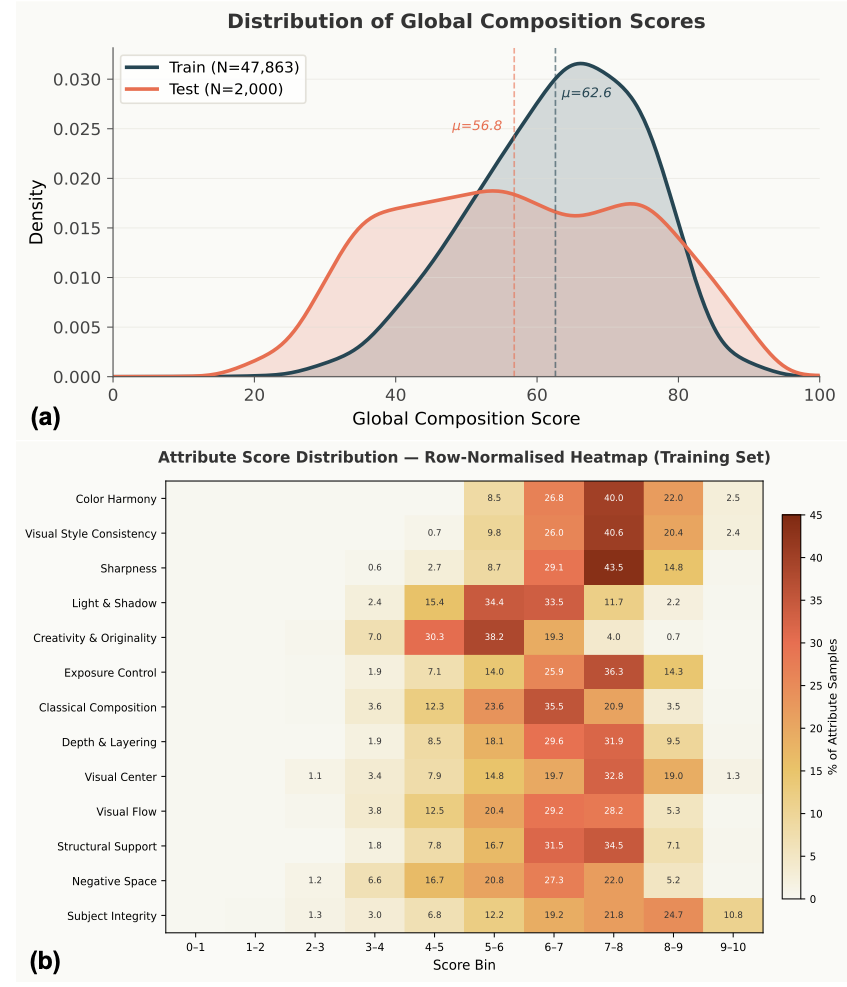}
\caption{Statistics of Track~1 composition annotations.
(a) Smoothed density distributions of global composition scores for the training and test sets. The test set shows a broader distribution and slightly lower average scores, indicating increased evaluation difficulty.
(b) Radar plot of mean scores across the 13 composition attributes. The variation across attributes reflects the heterogeneous characteristics of portrait composition and supports fine-grained evaluation beyond a single overall score.}
\label{fig:f2}
\end{figure}

\begin{table}[t]
\centering
\footnotesize
\setlength{\tabcolsep}{3.55pt}
\begin{tabular}{lcccccc}
\toprule
\textbf{Type} & \textbf{Samples} & \textbf{Mean} & \textbf{Std} & \textbf{Median} & \textbf{Range} & \textbf{\makecell[c]{Avg. Attr. \\ Length}} \\
\midrule
Train   & 4,500 & 2237.0 & 251.4 & 2214 & 1459--3389 & 172.0 \\
Test    & 500   & 2230.8 & 240.2 & 2215 & 1567--3163 & 171.2 \\
Overall & 5,000 & 2236.4 & 250.3 & 2214 & 1459--3389 & 171.9 \\
\bottomrule
\end{tabular}
\caption{Statistics of structured composition descriptions in PortraitCraft for Track~2. Mean, Std, Median, and Range summarize the distribution of total description length for each dataset split. Avg. Attr. Length denotes the average description length per composition attribute across the 13 predefined attributes. Lengths are measured in characters.}
\label{tab:track2_description_statistics}
\end{table}

\subsection{Annotation Framework}
The annotation framework in PortraitCraft combines expert involvement with model-assisted annotation. The annotation team includes professional designers, photographers, visual arts practitioners, and researchers with relevant expertise in aesthetics and visual analysis. Their experience ranges from more than three years to over ten years. Automatic models were used to improve efficiency, while expert review was used to maintain annotation quality. The annotation design is organized around the two benchmark tasks. For portrait composition understanding, the training set contains portrait images, overall composition scores, and attribute-level annotations for 13 composition attributes. Each attribute is first assigned a score and can be further mapped to three levels, namely high, medium, and low. In addition, the training data provide textual explanations for these scoring results, so that the supervision covers both structured judgments and their underlying rationale. In the test set, the model is required to predict the overall composition score, assign a high, medium, or low label to each of the 13 attributes, and answer one multiple-choice visual question based on the input image. For portrait composition generation, the dataset provides structured textual descriptions that are mainly centered on composition-related factors, such as subject arrangement, spatial organization, and visual emphasis. These descriptions are designed to connect explicit composition requirements with portrait image generation.

To improve scalability while preserving annotation quality, explanation text, VQA items, and structured composition descriptions were generated with advanced MLLMs and then reviewed and revised by experts. We also developed detailed annotation guidelines and standardized definitions for all 13 composition attributes. Annotators received training before the formal annotation process, and ambiguous cases were further reviewed during quality control. The training set was checked through sampled expert re-inspection, while the test set was fully reviewed to ensure higher reliability for benchmark evaluation.

\subsection{Dataset Statistics}

Track~1 contains 47,863 training samples and 2,000 test samples, resulting in a total of 49,863 annotated portrait images. Fig.~\ref{fig:f2}(a) shows the distribution of global composition scores for the two splits. The scores are mainly concentrated in the mid-to-high range, while the test set exhibits a lower mean score and a wider distribution than the training set, indicating increased evaluation difficulty and improved discriminative capability of the benchmark. Fig.~\ref{fig:f2}(b) illustrates the mean scores of the 13 composition attributes. The results show clear variations across attributes, suggesting that the predefined attribute space captures complementary aspects of portrait composition rather than a single underlying quality dimension. In addition, the explanation texts associated with attribute-level judgments remain stable in length across splits, providing consistent supervision for fine-grained composition understanding.

Table~\ref{tab:track2_description_statistics} summarizes the statistics of structured composition descriptions in Track~2. The average total description length exceeds 2,200 characters for all splits, indicating that the generation task is supported by detailed composition-aware textual supervision. The training and test sets show highly similar distributions in mean length, standard deviation, median length, and average attribute-level description length, which suggests stable annotation quality across splits. In addition, the average attribute-level description length is around 172 characters, showing that each of the 13 predefined composition attributes is supported by sufficiently detailed textual information. These statistics confirm that Track~2 provides consistent and fine-grained structured descriptions for composition-aware portrait generation.

\subsection{Usage Policy}
PortraitCraft is intended for academic research use only. The dataset is constructed from publicly available photographs collected from Unsplash and reorganized into a benchmark for portrait composition understanding and generation. Users of the dataset must comply with the original licensing terms of Unsplash. The dataset may not be used for commercial purposes or for applications involving identity recognition, surveillance, or other sensitive scenarios. We encourage responsible use of the dataset and recommend that users carefully consider ethical implications when applying the dataset to downstream tasks involving human subjects.
\section{Evaluation Protocol}
\label{sec:metrics}

\subsection{Evaluation Metrics for Portrait Composition Understanding}
Track~1 evaluates portrait composition understanding from three complementary aspects: overall composition score prediction, attribute-level composition judgment, and image-based visual question answering. The protocol is designed to assess whether a model can perform global composition assessment, recognize fine-grained composition attributes, and understand image details related to portrait structure. Specifically, models are required to predict an overall composition score for each image, assign one of three levels (high, medium, or low) to each of the 13 composition attributes, and answer one multiple-choice visual question based on the input image. Performance is evaluated by comparing model predictions with expert annotations for all three components. The final evaluation considers these components jointly to reflect the overall capability of structured portrait composition understanding.

\subsection{Evaluation Metrics for Portrait Composition Generation}
Track~2 evaluates portrait composition generation under structured composition constraints. The protocol focuses on whether a generated image matches the composition requirements expressed in the input description, rather than on general image realism or visual similarity alone. Specifically, participants are provided with structured composition descriptions and are required to generate corresponding portrait images. The submitted images are then analyzed by an internal model that extracts composition-related properties from the generated results. These properties are compared with the target textual specifications to measure composition alignment. The final evaluation is based on how well the generated image satisfies the intended composition structure and detailed composition requirements.

\section{Experiments}
\label{sec:exp}

\subsection{Experimental Setup and Baselines}
To validate the feasibility of the proposed benchmark and provide a practical reference baseline for future participants, we conducted a lightweight baseline experiment based on Qwen3-VL-4B~\cite{qwen3vl}. The purpose of this experiment is to establish an initial performance reference on PortraitCraft rather than to obtain fully optimized results. We adopted Qwen3-VL-4B~\cite{qwen3vl} as the base model and performed partial fine-tuning on the language model and MLP components while keeping the vision encoder frozen. The model was trained for 2 epochs with a learning rate of $1\times10^{-5}$, a batch size of 4, and 4 gradient accumulation steps. This setting enables efficient adaptation to the proposed tasks while keeping the training process computationally affordable. The resulting model serves as a simple and reproducible baseline to support preliminary evaluation on the PortraitCraft benchmark.

\subsection{Results on Portrait Composition Understanding}
Table~\ref{tab:track1_results} reports the baseline results on Track~1. We compare the zero-shot performance of Qwen3-VL-4B with the results obtained after partial fine-tuning on PortraitCraft. The fine-tuned model improves consistently over the zero-shot setting across all four metrics, which suggests that the proposed benchmark provides effective supervision for portrait composition understanding. The gains in SRCC and PLCC indicate better agreement with expert annotations in overall composition assessment, while the improvement in QA accuracy suggests stronger image-grounded reasoning ability after training on PortraitCraft. By contrast, the gain in attribute-level accuracy is relatively smaller, which indicates that fine-grained composition judgment across 13 attributes remains more challenging. Overall, these results show that the proposed dataset is useful for both global composition learning and fine-grained portrait understanding, while highlighting the remaining challenges in attribute-level composition judgment.

\begin{table}[t]
\centering
\footnotesize
\setlength{\tabcolsep}{6.5pt}
\begin{tabular}{lcccc}
\toprule
\textbf{Method} & \textbf{SRCC$\uparrow$} & \textbf{PLCC$\uparrow$} & \textbf{Level Acc$\uparrow$} & \textbf{QA Acc$\uparrow$} \\
\midrule
\makecell[l]{Qwen3-VL-4B\\(Zero-shot)} 
& 0.6726 & 0.6420 & 0.4290 & 0.7825 \\
\makecell[l]{Qwen3-VL-4B\\(Fine-tuned)} 
& \textbf{0.7778} & \textbf{0.7477} & \textbf{0.4592} & \textbf{0.8765} \\

\bottomrule
\end{tabular}
\caption{Baseline results on Track~1: Portrait Composition Understanding. All metrics are higher-is-better. Lightweight fine-tuning MLLMs on PortraitCraft improves performance over the zero-shot setting.}
\label{tab:track1_results}
\end{table}

\subsection{Results on Portrait Composition Generation}
\begin{figure}[t]
\centering
\includegraphics[width=\linewidth]{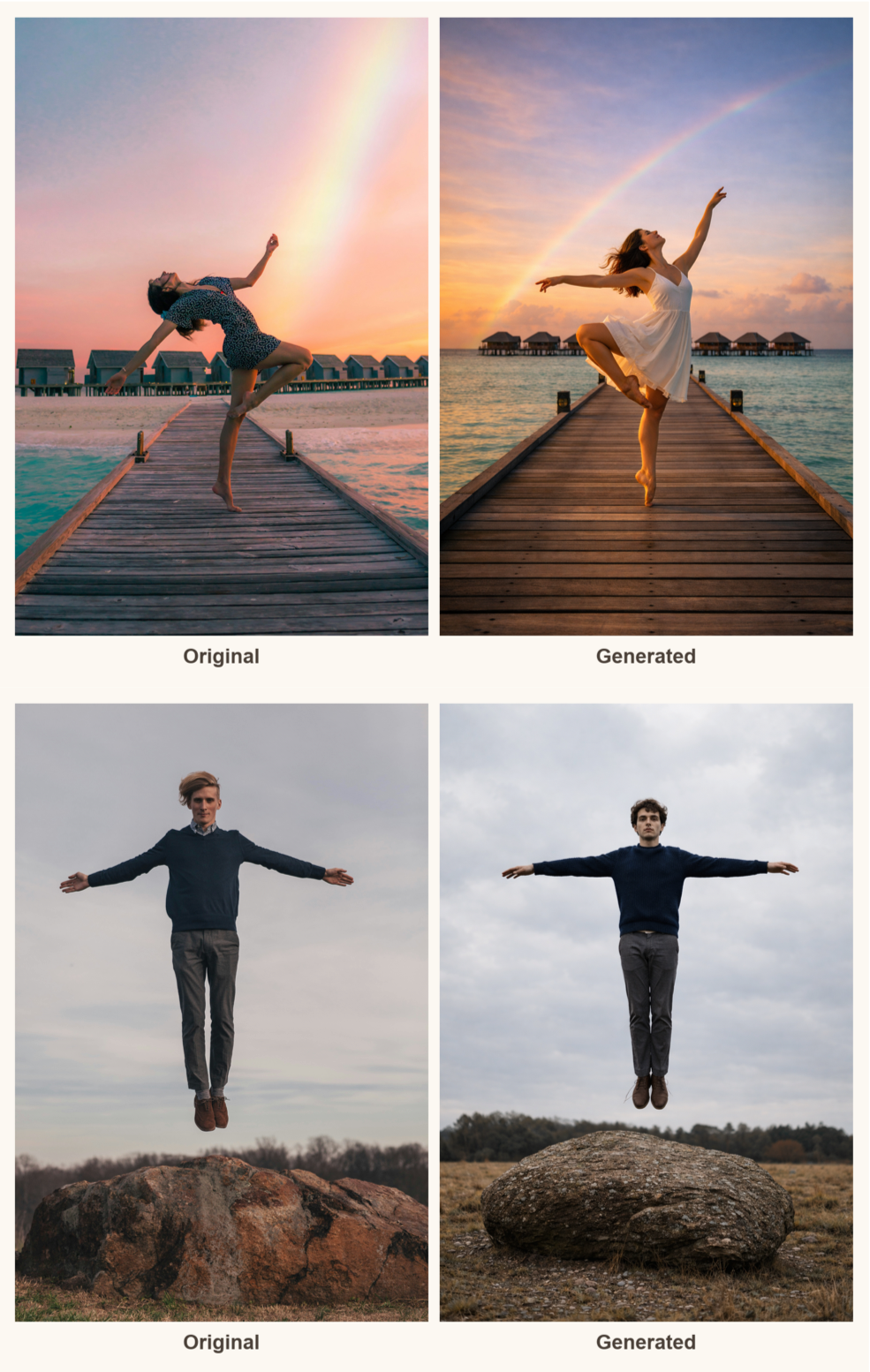}
\caption{Qualitative results on Track~2: Portrait Composition Generation. Two representative examples are shown. For each example, \textit{Original} denotes the reference portrait image associated with the structured composition description, and \textit{Generated} denotes the image produced by the model from the same composition specification. The comparison shows that the generated portraits follow key composition cues such as subject placement, spatial organization, and visual emphasis, demonstrating the feasibility of composition-aware portrait generation under the proposed benchmark setting.}
\label{fig:f3}
\end{figure}

We further evaluate Track 2 through qualitative generation results. After fine-tuning an existing MLLMs on the proposed benchmark, we observe that the model can generate visually plausible portrait images that follow the input composition descriptions to a reasonable extent. These results suggest that the portrait composition generation task is practically feasible and that the proposed dataset can support composition-aware generation research. Fig.~\ref{fig:f3} presents two representative examples. Each example includes the input composition description and the corresponding generated portrait image. Rather than emphasizing general image realism alone, these examples are intended to show whether the generated results reflect the key composition cues specified in the text. As shown in the figure, the generated portraits exhibit good alignment with several important composition-related factors, such as subject placement, spatial organization, visual emphasis, and overall scene arrangement. These qualitative results provide initial evidence that Track~2 is operational in practice and that existing models can serve as effective starting points for this task. At the same time, they also indicate that portrait composition generation remains a meaningful research problem, especially when more precise control over fine-grained composition structure is required.

\section{Conclusion}
\label{sec:conc}

In this paper, we introduced PortraitCraft, a benchmark for portrait composition understanding and generation. Unlike conventional aesthetic datasets that mainly focus on overall quality assessment, PortraitCraft formulates portrait composition as a structured benchmark problem and supports research on both composition analysis and composition-aware generation. By connecting these two directions within a unified framework, PortraitCraft provides a more complete setting for studying portrait composition beyond coarse aesthetic scoring. To support this benchmark, we constructed a portrait-centered dataset about 50,000 curated real images and developed an annotation framework that combines expert involvement with model-assisted annotation. The dataset provides multi-level supervision, including overall composition scores, fine-grained attribute annotations, explanation text, visual question answering pairs, and structured composition descriptions. These components enable PortraitCraft to support both fine-grained composition understanding and controllable portrait generation. 

We expect PortraitCraft to serve as a useful resource for future work on composition-aware vision and generation. In particular, the benchmark opens up new opportunities for studying fine-grained composition reasoning, interpretable portrait assessment, and structured control in portrait generation. We hope this work can encourage further progress toward more rigorous, explainable, and controllable research on portrait composition.
{
    \small
    \bibliographystyle{ieeenat_fullname}
    \bibliography{main}

@String(AAAI = {AAAI})

@article{artimuse,
  title={{Artimuse}: Fine-grained image aesthetics assessment with joint scoring and expert-level understanding},
  author={Cao, Shuo and Ma, Nan and Li, Jiayang and Li, Xiaohui and Shao, Lihao and Zhu, Kaiwen and Zhou, Yu and Pu, Yuandong and Wu, Jiarui and Wang, Jiaquan and others},
  journal={arXiv preprint arXiv:2507.14533},
  year={2025}
}

@inproceedings{ava,
  title={{AVA}: A video dataset of spatio-temporally localized atomic visual actions},
  author={Gu, Chunhui and Sun, Chen and Ross, David A. and Vondrick, Carl and Pantofaru, Caroline and Li, Yeqing and Vijayanarasimhan, Sudheendra and Toderici, George and Ricco, Susanna and Sukthankar, Rahul and others},
  booktitle={Proceedings of the IEEE Conference on Computer Vision and Pattern Recognition},
  pages={6047--6056},
  year={2018}
}

@inproceedings{aadb,
  title={Photo aesthetics ranking network with attributes and content adaptation},
  author={Kong, Shu and Shen, Xiaohui and Lin, Zhe and Mech, Radomir and Fowlkes, Charless},
  booktitle={Proceedings of the European Conference on Computer Vision},
  pages={662--679},
  year={2016}
}

@inproceedings{apddv2,
  title={{APDDv2}: Aesthetics of paintings and drawings dataset with artist labeled scores and comments},
  author={Jin, Xin and Qiao, Qianqian and Lu, Yi and Wang, Huaye and Huang, Heng and Gao, Shan and Liu, Jianfei and Li, Rui},
  booktitle={Advances in Neural Information Processing Systems},
  year={2024}
}

@inproceedings{c-3,
  title={{Science-t2i}: Addressing scientific illusions in image synthesis},
  author={Li, Jialuo and Chai, Wenhao and Fu, Xingyu and Xu, Haiyang and Xie, Saining},
  booktitle={Proceedings of the IEEE/CVF Conference on Computer Vision and Pattern Recognition},
  pages={2734--2744},
  year={2025}
}

@inproceedings{kacl,
  title={Contrastive knowledge-guided large language models for medical report generation},
  author={Sha, Yuyang and Pan, Hongxin and Meng, Weiyu and Li, Kefeng},
  booktitle={Medical Image Computing and Computer Assisted Intervention -- MICCAI},
  pages={111--120},
  year={2025}
}

@inproceedings{piq23,
  title={An image quality assessment dataset for portraits},
  author={Chahine, Nicolas and Calarasanu, Stefania and Garcia-Civiero, Davide and Cayla, Theo and Ferradans, Sira and Ponce, Jean},
  booktitle={Proceedings of the IEEE/CVF Conference on Computer Vision and Pattern Recognition},
  pages={9968--9978},
  year={2023}
}

@inproceedings{para,
  title={Personalized image aesthetics assessment with rich attributes},
  author={Yang, Yuzhe and Xu, Liwu and Li, Leida and Qie, Nan and Li, Yaqian and Zhang, Peng and Guo, Yandong},
  booktitle={Proceedings of the IEEE/CVF Conference on Computer Vision and Pattern Recognition},
  pages={19861--19869},
  year={2022}
}

@article{qwen3vl,
  title={{Qwen3-VL} technical report},
  author={Bai, Shuai and Cai, Yuxuan and Chen, Ruizhe and Chen, Keqin and Chen, Xionghui and Cheng, Zesen and Deng, Lianghao and Ding, Wei and Gao, Chang and Ge, Chunjiang and others},
  journal={arXiv preprint arXiv:2511.21631},
  year={2025}
}

@inproceedings{ase-1,
  title={Thinking image color aesthetics assessment: Models, datasets and benchmarks},
  author={He, Shuai and Ming, Anlong and Li, Yaqi and Sun, Jinyuan and Zheng, ShunTian and Ma, Huadong},
  booktitle={Proceedings of the IEEE/CVF International Conference on Computer Vision},
  pages={21838--21847},
  year={2023}
}

@article{mdd-llm,
  title={{MDD-LLM}: Towards accuracy large language models for major depressive disorder diagnosis},
  author={Sha, Yuyang and Pan, Hongxin and Xu, Wei and Meng, Weiyu and Luo, Gang and Du, Xinyu and Zhai, Xiaobing and Tong, Henry H. Y. and Shi, Caijuan and Li, Kefeng},
  journal={Journal of Affective Disorders},
  volume={388},
  pages={119774},
  year={2025},
  publisher={Elsevier}
}

@inproceedings{flickr,
  title={Personalized image aesthetics},
  author={Ren, Jian and Shen, Xiaohui and Lin, Zhe and Mech, Radomir and Foran, David J.},
  booktitle={Proceedings of the IEEE International Conference on Computer Vision},
  pages={638--647},
  year={2017}
}

@inproceedings{c-1,
  title={{Id-Sculpt}: {Id}-aware 3D head generation from single in-the-wild portrait image},
  author={Hao, Jinkun and Tang, Junshu and Zhang, Jiangning and Yi, Ran and Hong, Yijia and Li, Moran and Cao, Weijian and Wang, Yating and Wang, Chengjie and Ma, Lizhuang},
  booktitle={Proceedings of the AAAI Conference on Artificial Intelligence},
  volume={39},
  number={3},
  pages={3383--3391},
  year={2025}
}

@inproceedings{c-2,
  title={Can machines understand composition? Dataset and benchmark for photographic image composition embedding and understanding},
  author={Zhao, Zhaoran and Lu, Peng and Zhang, Anran and Li, Peipei and Li, Xia and Liu, Xuannan and Hu, Yang and Chen, Shiyi and Wang, Liwei and Guo, Wenhao},
  booktitle={Proceedings of the IEEE/CVF Conference on Computer Vision and Pattern Recognition},
  pages={14411--14421},
  year={2025}
}

@inproceedings{finecaption,
  title={{Finecaption}: Compositional image captioning focusing on wherever you want at any granularity},
  author={Hua, Hang and Liu, Qing and Zhang, Lingzhi and Shi, Jing and Kim, Soo Ye and Zhang, Zhifei and Wang, Yilin and Zhang, Jianming and Lin, Zhe and Luo, Jiebo},
  booktitle={Proceedings of the IEEE/CVF Conference on Computer Vision and Pattern Recognition},
  pages={24763--24773},
  year={2025}
}

@inproceedings{unireal,
  title={{Unireal}: Universal image generation and editing via learning real-world dynamics},
  author={Chen, Xi and Zhang, Zhifei and Zhang, He and Zhou, Yuqian and Kim, Soo Ye and Liu, Qing and Li, Yijun and Zhang, Jianming and Zhao, Nanxuan and Wang, Yilin and others},
  booktitle={Proceedings of the IEEE/CVF Conference on Computer Vision and Pattern Recognition},
  pages={12501--12511},
  year={2025}
}

@article{mdd-thinker,
  title={{MDD-thinker}: A reasoning-enhanced large language model for diagnosis of major depressive disorder},
  author={Sha, Yuyang and Pan, Hongxin and Luo, Gang and Shi, Caijuan and Chen, Wei and Wang, Jing and Li, Kefeng},
  journal={Journal of Affective Disorders},
  volume={403},
  pages={121405},
  year={2026},
  publisher={Elsevier}
}

@inproceedings{lapis,
  title={{LAPIS}: A novel dataset for personalized image aesthetic assessment},
  author={Maerten, Anne-Sofie and Chen, Li-Wei and De Winter, Stefanie and Bossens, Christophe and Wagemans, Johan},
  booktitle={Proceedings of the IEEE/CVF Conference on Computer Vision and Pattern Recognition},
  pages={6302--6311},
  year={2025}
}

@article{c-4,
  title={{Consistentid}: Portrait generation with multimodal fine-grained identity preserving},
  author={Huang, Jiehui and Dong, Xiao and Song, Wenhui and Chong, Zheng and Tang, Zhenchao and Zhou, Jun and Cheng, Yuhao and Chen, Long and Li, Hanhui and Yan, Yiqiang and others},
  journal={IEEE Transactions on Pattern Analysis and Machine Intelligence},
  year={2026},
  publisher={IEEE}
}

@inproceedings{c-5,
  title={{EchoShot}: Multi-shot portrait video generation},
  author={Wang, Jiahao and Sheng, Hualian and Cai, Sijia and Zhang, Weizhan and Yan, Caixia and Feng, Yachuang and Deng, Bing and Ye, Jieping},
  booktitle={Advances in Neural Information Processing Systems},
  year={2025}
}

@inproceedings{c-6,
  title={{HiFi-Portrait}: Zero-shot identity-preserved portrait generation with high-fidelity multi-face fusion},
  author={Xu, Yifang and Zhai, Benxiang and Sun, Yunzhuo and Li, Ming and Li, Yang and Du, Sidan},
  booktitle={Proceedings of the IEEE/CVF Conference on Computer Vision and Pattern Recognition},
  pages={5625--5635},
  year={2025}
}

@article{llava,
  title={{LLaVA}: Visual instruction tuning},
  author={Liu, Haotian and Li, Chunyuan and Wu, Qingyang and Lee, Yong Jae},
  journal={Advances in Neural Information Processing Systems},
  volume={36},
  pages={34892--34916},
  year={2023}
}

@article{vlm,
  title={Vision-language models for vision tasks: A survey},
  author={Zhang, Jingyi and Huang, Jiaxing and Jin, Sheng and Lu, Shijian},
  journal={IEEE Transactions on Pattern Analysis and Machine Intelligence},
  volume={46},
  number={8},
  pages={5625--5644},
  year={2024},
  publisher={IEEE}
}

@inproceedings{c-7,
  title={{Argus}: Vision-centric reasoning with grounded chain-of-thought},
  author={Man, Yunze and Huang, De-An and Liu, Guilin and Sheng, Shiwei and Liu, Shilong and Gui, Liang-Yan and Kautz, Jan and Wang, Yu-Xiong and Yu, Zhiding},
  booktitle={Proceedings of the IEEE/CVF Conference on Computer Vision and Pattern Recognition},
  pages={14268--14280},
  year={2025}
}

@inproceedings{c-8,
  title={{Hyperlora}: Parameter-efficient adaptive generation for portrait synthesis},
  author={Li, Mengtian and Chen, Jinshu and Feng, Wanquan and Li, Bingchuan and Dai, Fei and Zhao, Songtao and He, Qian},
  booktitle={Proceedings of the IEEE/CVF Conference on Computer Vision and Pattern Recognition},
  pages={13114--13123},
  year={2025}
}
}


\end{document}